\documentclass[11pt]{article}
\usepackage[a4paper,margin=1in]{geometry}
\usepackage[T1]{fontenc}
\usepackage[utf8]{inputenc}
\usepackage{lmodern}
\usepackage{graphicx}
\usepackage{amsmath,amssymb}
\usepackage{booktabs,array,longtable}
\usepackage{microtype}
\usepackage{url}
\usepackage[hidelinks]{hyperref}
\usepackage[round,authoryear]{natbib}
\usepackage{caption}
\usepackage{float}
\title{Quantitative Measurement of Pavement Distress based on Vehicle Camera and Perspective Correction}
\author{Sike Wang$^{1}$ \and Yizhou Lin$^{1}$ \and Shuo Li$^{2}$ \and Yi Jiang$^{1}$ \and Chengcheng Tao$^{1}$\\[0.6em]
\small $^1$Bowen School of Construction, Purdue University, West Lafayette, IN, USA\\
\small $^2$Indiana Department of Transportation, Division of Research and Development, West Lafayette, IN, USA\\
}

\date{}
\begin{document}
\maketitle
\begin{abstract}
Using image data is fast becoming a key method for
highway agencies to automatically evaluate pavement conditions, while
most of the research focuses on developing more accurate algorithms
instead of obtaining the quantitative length or area of pavement
distress. Although the corresponding values can be obtained by
calibrating the cameras in the field, this is not feasible for large
volumes of onboard camera data from different vehicles. The purpose of
this study is to obtain quantitative measurements of road surface damage
directly from in-vehicle camera data, without the need for on-site
calibration of the cameras.
Based on a manually selected pavement image, the
region of interest in a quadrilateral is obtained from the lane
detection, which is utilized to construct a perspective matrix for
perspective correction of all images. Then, the size correction is
conducted based on the proportions of the road markings in the image and
the specified scale, which can also determine the actual width of the
selected region of interest. Finally, the adjusted images are fed into
object detection algorithms to obtain quantitative detection results.
The proposed methodology could efficiently conduct
perspective correction for all images collected by the same vehicle
camera with minimal manual effort. The quantitative pavement patch
length measured by the proposed method exhibits the same trend as the
pavement friction data.
 This study calibrates vehicle cameras using only image
data, without the need for additional experiments, thereby enabling the
quantitative detection of pavement distress, which is an aspect that
previous studies have generally overlooked. 
This research provides a practical
method to process a large amount of vehicle camera data to achieve
quantitative measurement of pavement distress, especially with the
increased number of connected vehicles that provide massive amounts of
uncalibrated data.
\end{abstract}

\section{Introduction}\label{introduction}

Since 1980s, imaging techniques have become a practical way to evaluate
pavement conditions, which prevent extensive manual labor and traffic
interruption \citep{Fundakowski1991}. However, early data processing
methods were unable to automatically detect pavement damage, so manual
inspection of images was still required. With the development of
computer vision technologies, some automatic data processing methods
were introduced into pavement inspection, such as OTSU, which is
typical threshold method used to segment pavement distress \citep{Li2008,Akagic2018}, and edge detection methods that can determine
the edge of distress \citep{Zhao2010,NhatDuc2018}. Although
these methods achieve automatic survey of pavement, their effects are
significantly influenced by the environment, especially when photos are
affected by lighting or road surface stains, it is difficult to achieve
accurate detection results \citep{Mulyanto2025}.

Recent developments in deep learning have led to increased interest in
developing robust and automatic pavement inspection methods. Based
on different tasks, the methods can be classified into three categories:
classification, object detection, and segmentation. The task of
classification is determining the different pavement conditions from
images, which only provides classification information without any
location and other detailed measurement \citep{Doycheva2017,Gopalakrishnan2017}. In addition to image classification, object
detection is another mainstream method for detecting pavement distress.
Different from image classification, object detection algorithms can
output bounding boxes for entities within an image for subsequent
processing. Therefore, object detection methods could provide more
specific location information for pavement management \citep{Maeda2018}, and rough distress measurements \citep{WangS2024}. In
addition, another widely used detection algorithm is segmentation, which
can identify the specific location and shape of pavement distress,
enabling precise measurement of pixels \citep{Kheradmandi2022}.

Although more and more highway agencies have equipped with professional
pavement inspection vehicle that can collect vertical-view images,
using the front-view images from the low-cost vehicle cameras is a
practical approach in inspecting pavement damage \citep{ZhangA2024}, especially the increasing connected vehicles provide a vast
collection of unlabeled road images. As introduced in previous chapter,
different pavement distress can be detected from image data, such as
patches \citep{Ning2024}, cracks \citep{Ning2024,Yang2024}
and potholes \citep{Chougule2023}. Although the segmentation
algorithms based on deep learning were widely used to extract the
detailed shapes of the pavement distress, such as crack pixels \citep{Yang2024} \citep{Ji2025} and the area of potholes \citep{Masihullah2021}. The distress measurement by segmentation algorithms requires a
high-computation platform and a large amount of segmented data for
training and running, which is costly and time-consuming, especially for
data labeling. Also, research on the subject has been mostly restricted
to improving the accuracy and speed of algorithms instead of obtaining
the actual measurement values of distress, such as the length of cracks,
the actual area of potholes. The segmented results from the previous
lack significant engineering value and cannot be integrated or
cross-analyzed with other data with absolute value, such as pavement
texture data, traffic volume.

This research aims to explore the possibility of using vehicle camera
data to quantitatively measure pavement distress without camera
calibration. Based on the lane detection results, the region of interest
is automatically selected to calculate homography matrix for perspective
adjustment. Then, the actual width of the region of interest is
determined by the size of broken line in the image, which is used for
size adjustment of images. Finally, the adjusted images are detected by
object detection algorithms to obtain the quantitative measurements of
pavement distress. It is hoped that this research will contribute to
increasing the use of vast vehicle images for quantitative pavement
inspection without utilizing expensive professional equipment.

\section{Methods}\label{methods}

\subsection{Pavement Patch Measurement from Front-View
Image}\label{pavement-patch-measurement-from-front-view-image}

\subsubsection{Perspective Image
Adjustment}\label{perspective-image-adjustment}

\begin{figure}[h]
\centering
\includegraphics[width=\linewidth,height=0.66\textheight,keepaspectratio]{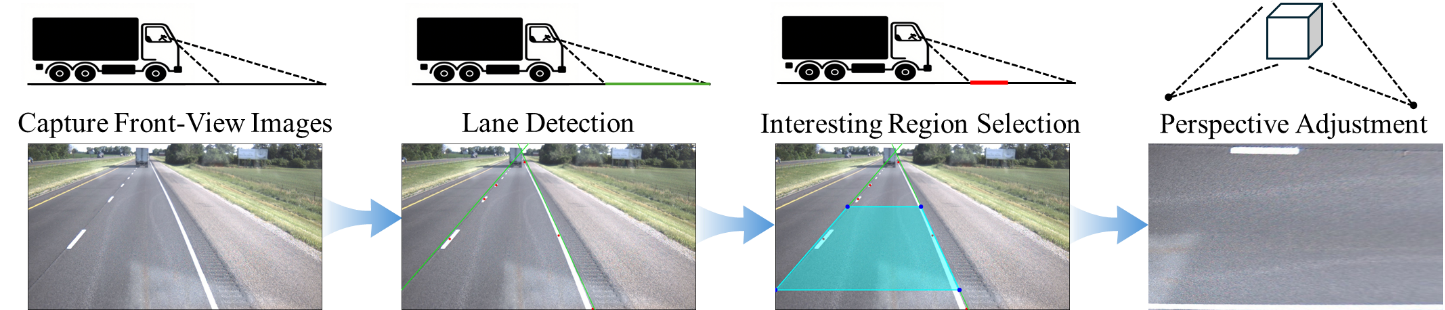}
\caption{Perspective adjustment of front-view images}\label{fig:1}
\end{figure}

Compared to the vertical images collected by professional inspection
vehicles, front-view images were generally captured at a fixed angle to
the pavement, which cannot be directly used to measure the distress
quantitatively. However, it is essential to get the quantitative value
of pavement damage for the friction model improvement and engineering
practice. Therefore, a view adjustment method is introduced in this
section to transfer the front-view images to the vertical view, as shown
in \ref{fig:1}. The adjustment consisted of automatic lane detection,
selecting the interesting regions and adjusting the perspective.

\begin{figure}[h]
\centering
\includegraphics[width=0.6\linewidth,height=0.66\textheight,keepaspectratio]{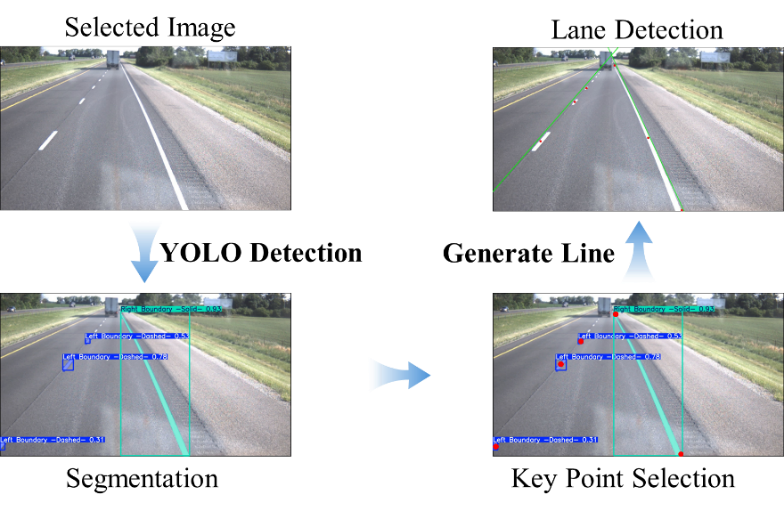}
\caption{Lane detection}\label{fig:2}
\end{figure}

After acquiring the front-view images, YOLOv11s was used to detect the
lane on the selected clean image. To achieve a satisfying detection
effect, the YOLOv11s model was trained in a public lane segmentation
dataset \citep{TobiasPrice2024} and used
pretrained weights provided by Ultralytics. The details of YOLOv11s
training will be introduced in Section \ref{crack-patch-length-measurement-based-on-yolov11}. 
Figure \ref{fig:2} shows the lane detection process which consisted of two parts: first, the YOLO model
recognized each lane line, including the separate lines on the left and
the continuous line on the right; then, the lane edge was generated from
the segmentation results. For the separate lines, the center points of
each detected line were connected by straight lines to form the lane
edges. For the continuous line, the diagonal line of the bounding box
was considered as the edge of the lane. Although considering lane edges
as straight lines might cause some error, the interesting region in this
study is close to the vehicle so this error is acceptable. The final
edge line will be shifted outward by 10 pixels to ensure it encompasses
the entire lane marking.

Due to the complexity of traffic and the scope of this research, the
lane region, which was close to the vehicle, was cropped in the process
as shown in Figure \ref{fig:1}. During the cropping process, the algorithm
considers this region as a trapezoid with parallel upper and lower
edges. In the research, the distance between the upper and lower edges
was set as half of the height of the image. This parameter can be
manually adjusted by the actual scenario.

The selection of the interesting region formed four corner points, which
were also used to calculate the homography matrix. Then, the homography
matrix was used to adjust all the points in the interesting region to
the vertical view as the last picture in Figure 1. In the research, the
target image size is set as 800 × 800, which means the transformation by
the homography matrix could convert the trapezoidal region of interest
in the original image to a square. It is important to note that in
perspective transformations, only the selected interesting region is
retained in the result, while the rest of the image is cropped.

Because the capture angle of the front-view camera is usually fixed for
a single vehicle, the homography matrix only needs to be calculated
once, and all the front-view images collected from the same camera can
be adjusted by this matrix. In the whole process, the operator only
needs to input an image and set the hight of the interesting area, then
the software would automatically adjust the view and cut the interesting
area.

\subsubsection{Image Size Adjustment}\label{image-size-adjustment}

\begin{figure}[htbp]
\centering
\includegraphics[width=\linewidth,height=0.66\textheight,keepaspectratio]{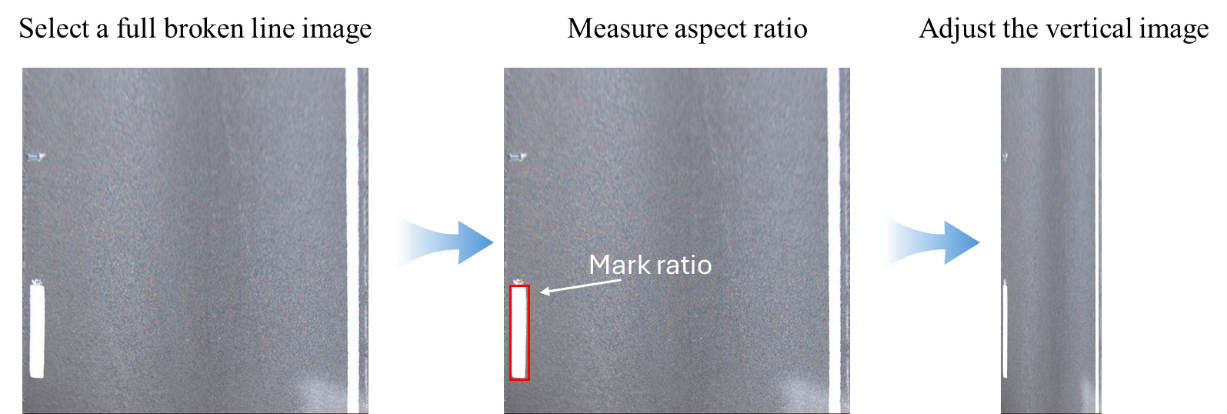}
\caption{Size adjustment of vertical pavement images}\label{fig:3}
\end{figure}

After obtaining the vertical perspective images, we are still unable to
directly acquire the quantitative dimensions of the damage, as the scale
between the photographs and the actual object remains unknown. To solve
this problem, an automatic size adjustment algorithm is proposed in this
section based on the pavement marking. According to the standards and
regulations for roadway pavement markings, the common broken line for a
highway is 10 feet long and 4-6 inches wide \citep{FHWA2024}. Based on this knowledge, we can adjust the
size of the images to conform to the actual scale. Figure \ref{fig:3} demonstrates
the workflow of the image size adjustment. First, a vertical image with
full broken lines is selected as the detected image. Then, the rectangle
boundary of the broken line is automatically detected, and the aspect
ratio of the broken line is calculated by the detected boundary. Last,
the height of the vertical image is adjusted based on the detected and
actual aspect marking ratios as shown in Equation \ref{eq:height}, and the width of
the vertical image is kept the same. The adjustment value can also be
used to adjust other vertical images from the same camera as Section
\ref{perspective-image-adjustment}, which means that we only need to manually select one image in
order to process all images captured by a single front-view camera.

\begin{equation}\label{eq:height} h^{\prime}=h\frac{R_t}{R_i} \end{equation}

where, \(h'\) is the adjusted height, \(h\) is the original height,
\(R_{t}\) is the actual aspect ratio of broken line (We adopt 20 here,
with the marking width assumed to be 6 inches.), \(R_{i}\) is the
measured aspect ratio of the broken line.

\begin{figure}[htbp]
\centering
\includegraphics[width=\linewidth,height=0.66\textheight,keepaspectratio]{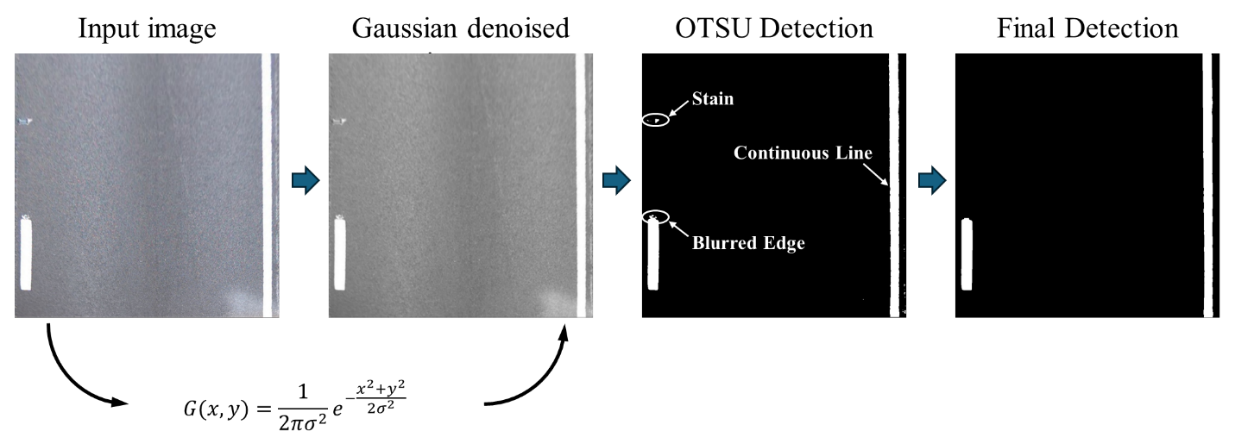}
\caption{Detection algorithm of pavement broken lines}\label{fig:4}
\end{figure}

Different from Section \ref{perspective-image-adjustment}, the marking region is automatically
detected in the size adjustment process, which is faster and more
accurate. The schema of the marking region detection algorithm is shown
in Figure \ref{fig:4}. The input image is adjusted by Gaussian Blur \citep{OpenCV2025}
to remove noise in the picture. Then, given the fact that there are
significant grayscale differences between road markings and the
pavement, the OTSU automatic thresholding method is employed to segment
and extract the broken line regions. However, due to the complex
background in the pavement images, the region extracted from OTSU
threshold might contain a lot of errors and disconnections. The erosion
and dilation are implemented in the algorithm to eliminate the burred
boundary and connect some detection missing separately. Last, the
extracted regions are fitted into rectangles, and the aspect ratio is
calculated. Furthermore, the area threshold is adopted in the algorithm
to mitigate the influence of small objects with high grayscale values
such as white stains and stones and the presence of continuous lines.
So, the final calculation of aspect ratios would only contain broken
lines to ensure a correct size adjustment.

\subsubsection{Crack-Patch Length Measurement based on
YOLOv11}\label{crack-patch-length-measurement-based-on-yolov11}

\begin{figure}[h]
\centering
\includegraphics[width=0.6\linewidth,height=0.66\textheight,keepaspectratio]{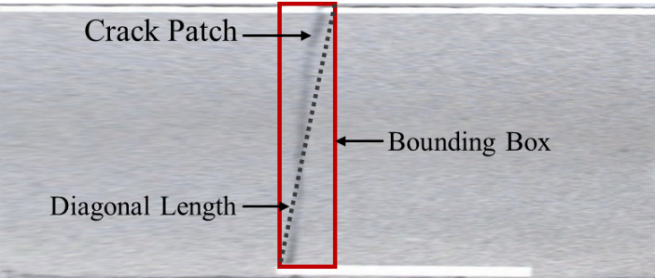}
\caption{Measurement of crack-patch length}\label{fig:5}
\end{figure}

Crack patches are one of the most common distresses on pavement, which
usually exhibit a linear configuration. Instead of using time-consuming
segmentation algorithms, we utilize a fast object algorithm to estimate
the diagonal length of crack-patch bonding box, whose effect has been
proven by linear crack measurement \citep{WangS2024}. 
Figure \ref{fig:5} demonstrates the length calculation by using the bounding box detected
from the object detection algorithm. Because the view and size of the
front-view images have been adjusted in Section \ref{perspective-image-adjustment} and \ref{image-size-adjustment}, the
diagonal length obtained from the detection can be directly converted
into actual length, rather than the pixel number measured by general
segmentation methods. The length measurement algorithm developed in this
research can automatically calculate all the diagonal lengths in a
single image and output the sum of all the measured lengths.

To obtain the bounding box of the crack patch, YOLOv11 is adopted as the
object detection algorithm in this study. YOLOv11 is the official
version of You Only Look Once (YOLO) algorithm, which is also the
state-of-art object detection algorithm with high detection speed
\citep{Ultralytics2025}. YOLOv11 keeps the basic structure from the previous
versions and introduces new layers to increase its performance.

\begin{figure}[h]
\centering
\includegraphics[width=\linewidth,height=0.66\textheight,keepaspectratio]{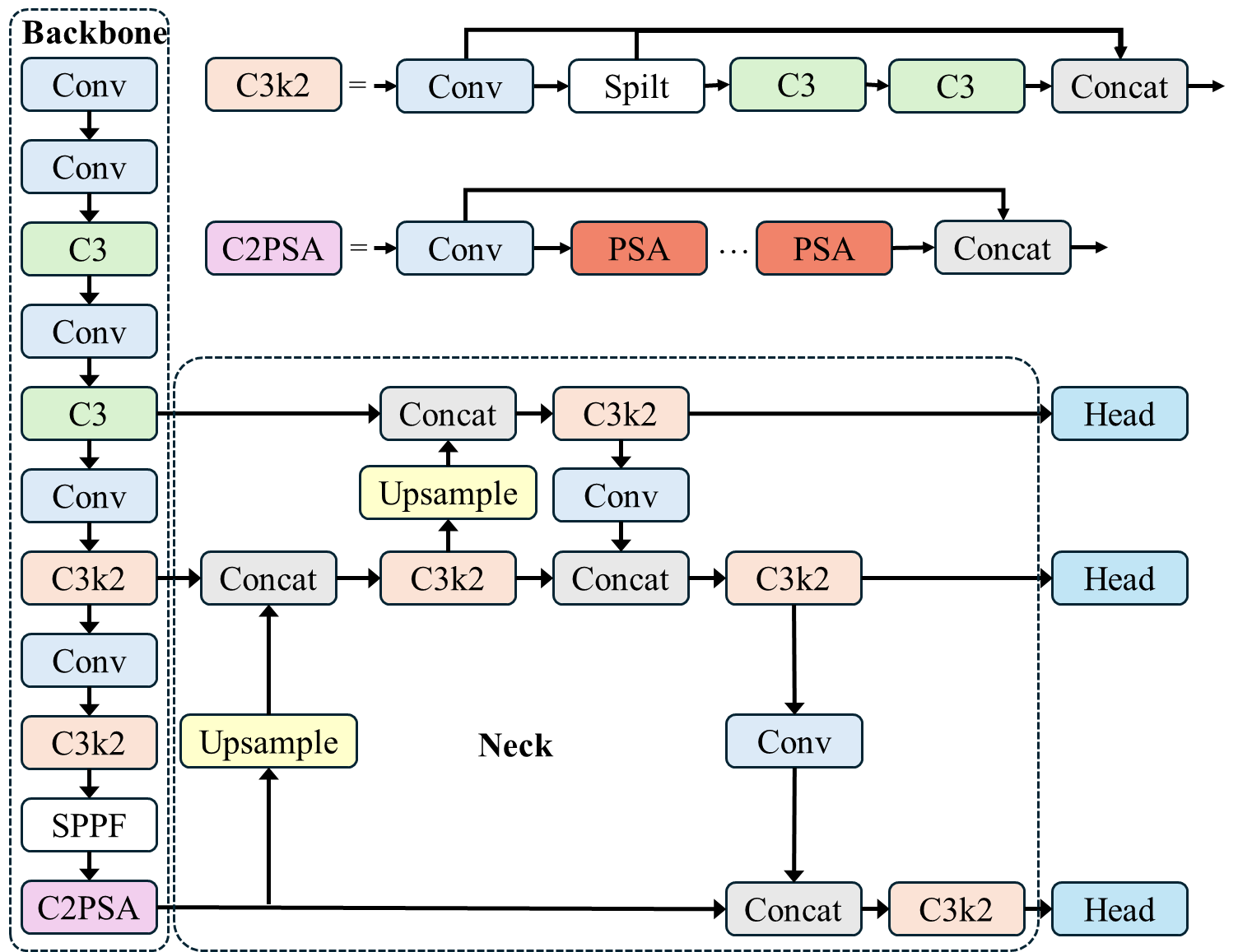}
\caption{Structure of YOLOv11}\label{fig:6}
\end{figure}

The component of YOLOv11 can be divided into three major parts as shown
in Figure \ref{fig:6}: (1) Backbone, (2) Neck, and (3) Head. Backbone is used to
interpret visual data by transforming raw image inputs into meaningful
feature representations. Neck is a crucial intermediate component
between Backbone and Head, which could merge the features from the
different layers in Backbone and create more various features. Head is
the task-specific component that takes processed features from the Neck
(or Backbone) and generates final predictions. YOLO v11 has been
optimized in all three main components compared to its previous
algorithms. In Backbone, a new convolutional layer, C3k2 was designed to
replace the original large convolutional~kernels with the stacked small
kernels. Also, the small kernel used in C3k2 is the deformable
convolutional kernels \citep{Dai2017} which provide a better
extraction of complex objects. A new cross-stage layer C2PSA was
developed for better extracting high-dimensional features. C2PSA adopted
the structure of CSPNet \citep{WangC2019} to separate the input
features to two parts. The first part is directly transmitted to output,
and other part is processed by Pyramid Squeeze Attention \citep{ZhangH2021} which introduces the attention mechanism and enhances the
model\textquotesingle s perception of object details. For Head, YOLOv11
employs depthwise separable convolution \citep{Chollet2017} to enhance the
model\textquotesingle s ability to detect complex objects. Introduction
of these techniques, coupled with the inherent superior performance of
YOLO, enables YOLO v11 to significantly enhance detection accuracy while
sustaining high detection speeds.

\section{\texorpdfstring{Experiments and Analysis
}{Experiments and Analysis }}\label{experiments-and-analysis}

First, the front-view images are adjusted to the vertical view, and
their sizes are corrected based on the size of the broken line in the
image. Then, an object detection model based on YOLOv11 is trained to
detect the region of crack patches. In the training of YOLOv11, we
adopted a refined method to improve the accuracy of YOLOv11, and
YOLOv11models with different sizes are compared to select a proper
model. Last, all the adjusted images are imported into the detection
model to obtain the bounding boxes of crack patches, and the diagonal
lengths of all the detected bounding boxes are automatically measured by
the developed algorithm.

\subsection{Adjustment of Front-view
Images}\label{adjustment-of-front-view-images}

\begin{figure}[h]
\centering
\includegraphics[width=\linewidth,height=0.66\textheight,keepaspectratio]{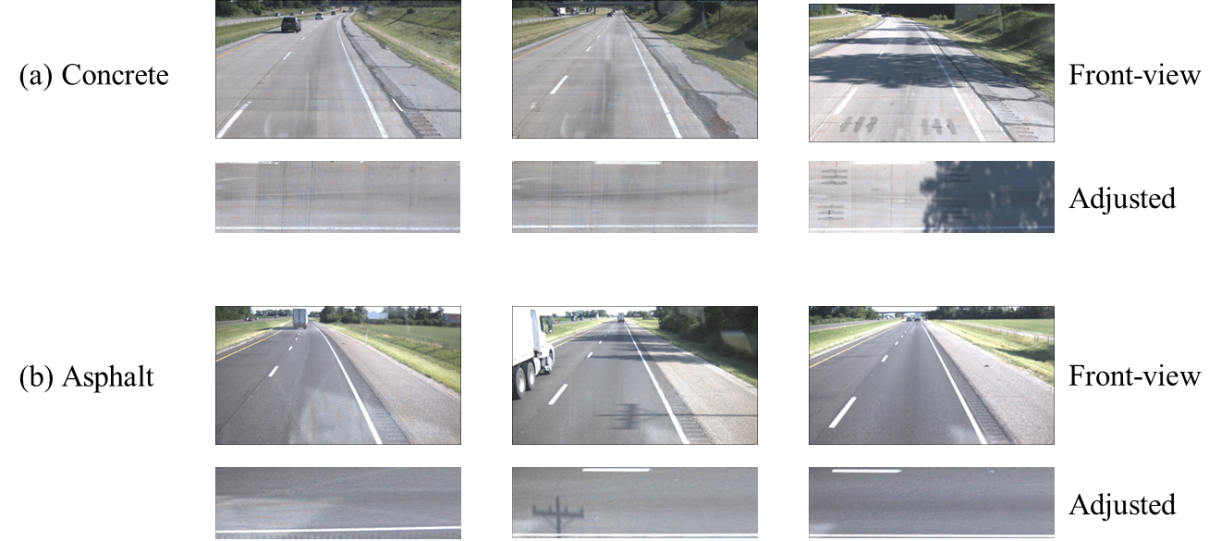}
\caption{SCRIM images adjustment}\label{fig:7}
\end{figure}

Figure \ref{fig:7} shows some samples from the front-view images collected by the
SCRIM test and their final adjustment results. The captured images
include both asphalt and cement pavement surfaces, with other vehicles
prominently featured in the photographs. However, the proposed
adjustment method requires selecting an interesting region manually,
which can eliminate the influence of the surrounding environment. In
this study, a front-view image with a full broken line was selected by
the operator. The selected image was captured on the asphalt pavement
because the black background provided by the asphalt surface creates a
stark contrast with the white road markings, which enhances the accuracy
of pavement marking extraction. The measured ratio of the broken line
from the selected image is 9.06, and the target . Also, the final
adjusted photo shows a covered area with a width of 150 inches (3.81 m)
and a length of 332 inches (8.43 m), which is calculated by the width of
the broken line. If photos with different ranges are required,
proportional cropping can be performed to retain the areas of greatest
interest.

\subsection{YOLOv11 Training for Patch
Detection}\label{yolov11-training-for-patch-detection}

\begin{figure}[h]
\centering
\includegraphics[width=\linewidth,height=0.66\textheight,keepaspectratio]{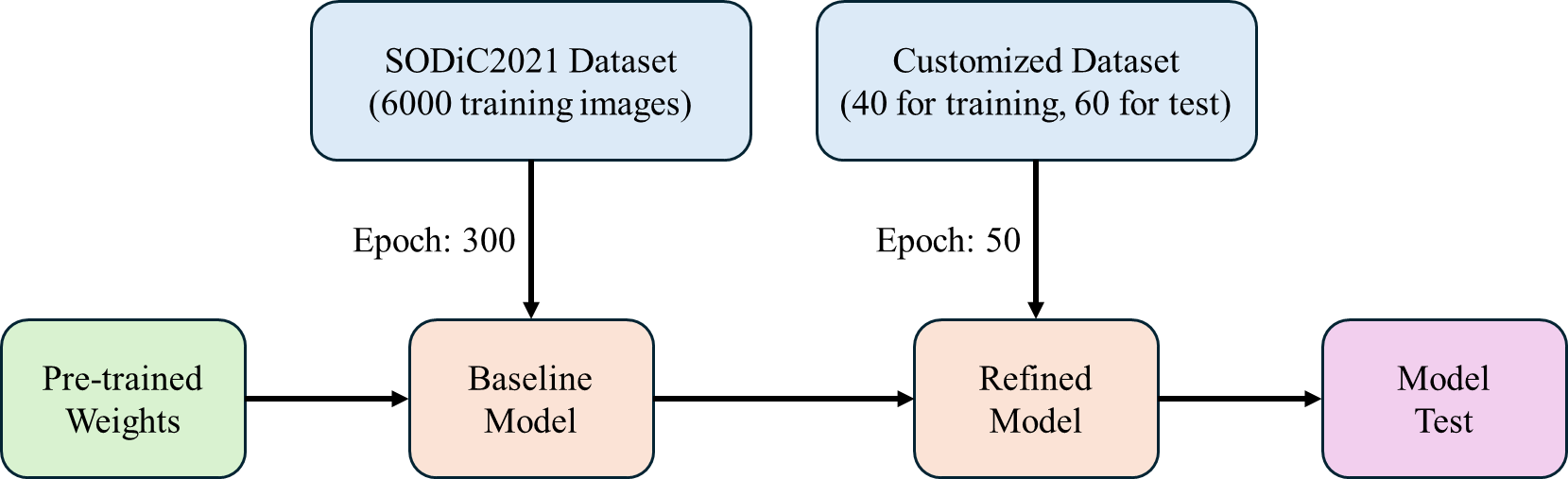}
\caption{YOLOv11 refined training}\label{fig:8}
\end{figure}

YOLOv11 is an object detection algorithm based on a deep neural network,
which needs a large amount of data to train. However, data about the
crack patch of pavement is extremely limited, and most of the images
were collected in front view. Therefore, we adopted a refined training
method \citep{WangS2024} as shown in Figure \ref{fig:8} to achieve an
acceptable detection accuracy. YOLOv11 is first trained in a large
pavement image dataset to get a baseline model. Then, the baseline model
is refined by a small, customized dataset that contains the images
collected in objective engineering. The large dataset used in the study
is the dataset provided by SODiC 2021 (PaddlePaddle, n.d.), which
contains eight kinds of pavement distress, including cracks, manholes,
nets, potholes, crack patches, net patches, and pothole patches. All the
images are front-view, and the total number of the training dataset is
6000. The customized dataset is built from 100 images randomly selected
from the adjusted SCRIM images. Meanwhile, the crack patches in the
pavement images were also annotated manually. To accelerate and improve
the training process, the pre-trained weights of YOLOv11 on COCO dataset
\citep{Lin2015} were used at the beginning of the training process.

We chose three different sizes of YOLOv11 (YOLOv11n, YOLOv11s, and
YOLOv11m) and two other versions of YOLO (YOLOv5s and YOLOv12s) to find
the best detection model. All the models were trained in the same
parameter setting as listed in Table \ref{tab:training} to ensure a fair comparison. And
the performance of each trained model was tested in the validation set
from the customized dataset.

\begin{table}[h]
\centering\caption{Parameter settings for model training}\label{tab:training}
\begin{tabular}{lr}\toprule
Parameter & Value \\ \midrule
Training Epoch (Baseline) & 300 \\
Early Stop (Baseline) & 50 \\
Training Epoch (Refine) & 50 \\
Early Stop (Refine) & 10 \\
Optimizer & SGD \\
Learning Rate & 0.01 \\
Batch Size & 32 \\
Input Size & 640 \\
\bottomrule\end{tabular}
\end{table}

\begin{figure}[h]
\centering
\includegraphics[width=\linewidth,height=0.66\textheight,keepaspectratio]{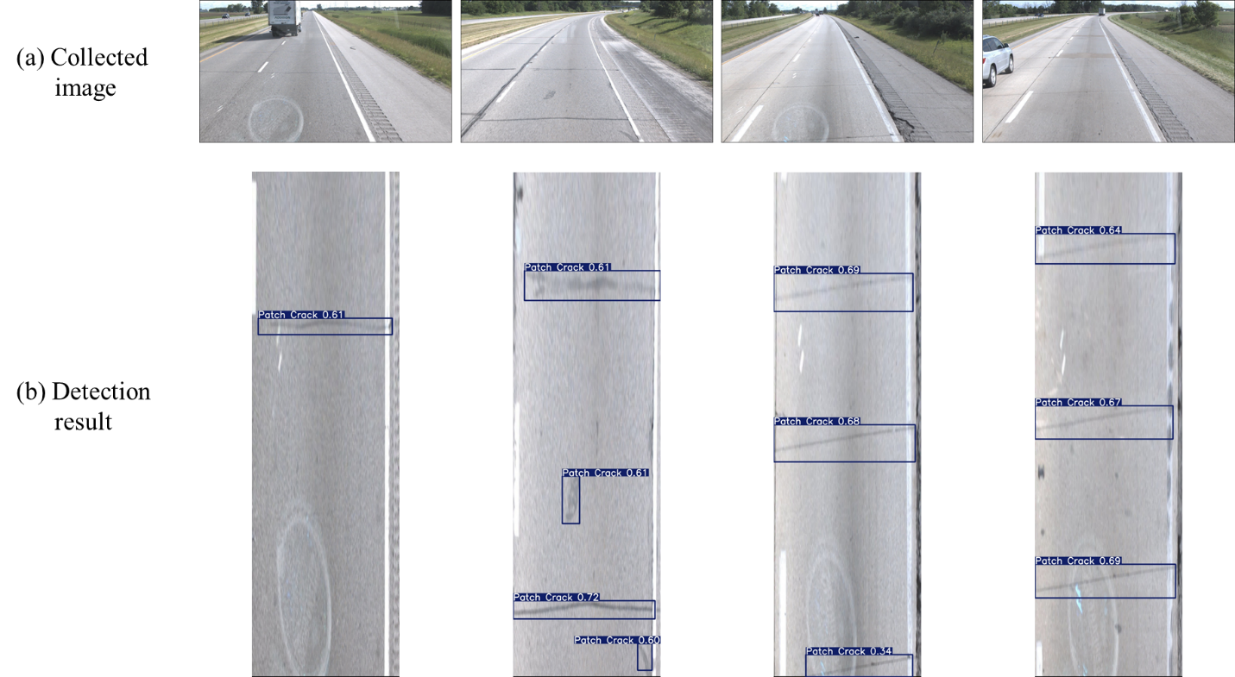}
\caption{Detection samples from the trained YOLOv11s}\label{fig:9}
\end{figure}

Table \ref{tab:models} compares the performance of the different models
including the model comparison before and after the refined training.
The accuracy of all models has been significantly enhanced after the
completion of refined training, which demonstrates the efficacy of the
training method. It noticed that the model without refined training
almost has no detection ability in the field test. Among all the models,
the refined YOLOv11m achieved the highest accuracy, while YOLOv11n
demonstrated the best speed performance. However, considering both
accuracy and speed, we selected YOLOv11s as the final detection model
due to its balanced performance. Figure \ref{fig:9} shows some result samples by
using the trained YOLOv11s to detect the crack patches. The final
detection results demonstrate a high level of accuracy, with effective
detection capabilities for both cement and asphalt road surfaces.

\begin{table}[htbp]
\centering\caption{Model test results}\label{tab:models}
\scriptsize\setlength{\tabcolsep}{3.6pt}
\begin{tabular}{lrrrrr}\toprule
Models & P & R & mAP@0.5 & Parameters & Speed (ms/image) \\ \midrule
YOLOv11n (With refine) & 0.873 & 0.783 & 0.826 & 2583712 & 2.7 \\
YOLOv11n (No refine) & 0.800 & 0.031 & 0.413 & 2583712 & 2.4 \\
YOLOv11s (With refine) & 0.897 & 0.813 & 0.881 & 9415896 & 3.1 \\
YOLOv11s (No refine) & 0.967 & 0.225 & 0.599 & 9415896 & 2.8 \\
YOLOv11m (With refine) & 0.940 & 0.845 & 0.883 & 20036200 & 5.0 \\
YOLOv11m (No refine) & 1.000 & 0.023 & 0.512 & 20036200 & 4.8 \\
YOLOv12s (With refine) & 0.907 & 0.752 & 0.822 & 9233976 & 5.4 \\
YOLOv12s (No refine) & 0.000 & 0.000 & 0.000 & 9233976 & 5.6 \\
YOLOv12m (With refine) & 0.871 & 0.822 & 0.837 & 20111080 & 6.2 \\
YOLOv12m (No refine) & 0.000 & 0.000 & 0.000 & 20111080 & 8.8 \\
YOLOv5s (With refine) & 0.918 & 0.690 & 0.815 & 9114632 & 2.9 \\
YOLOv5s (No refine) & 0.938 & 0.116 & 0.525 & 9114632 & 2.8 \\
\bottomrule\end{tabular}
\end{table}

\subsection{Length Measurement Results}

\begin{figure}[htbp]
\centering
\includegraphics[width=\linewidth,height=0.66\textheight,keepaspectratio]{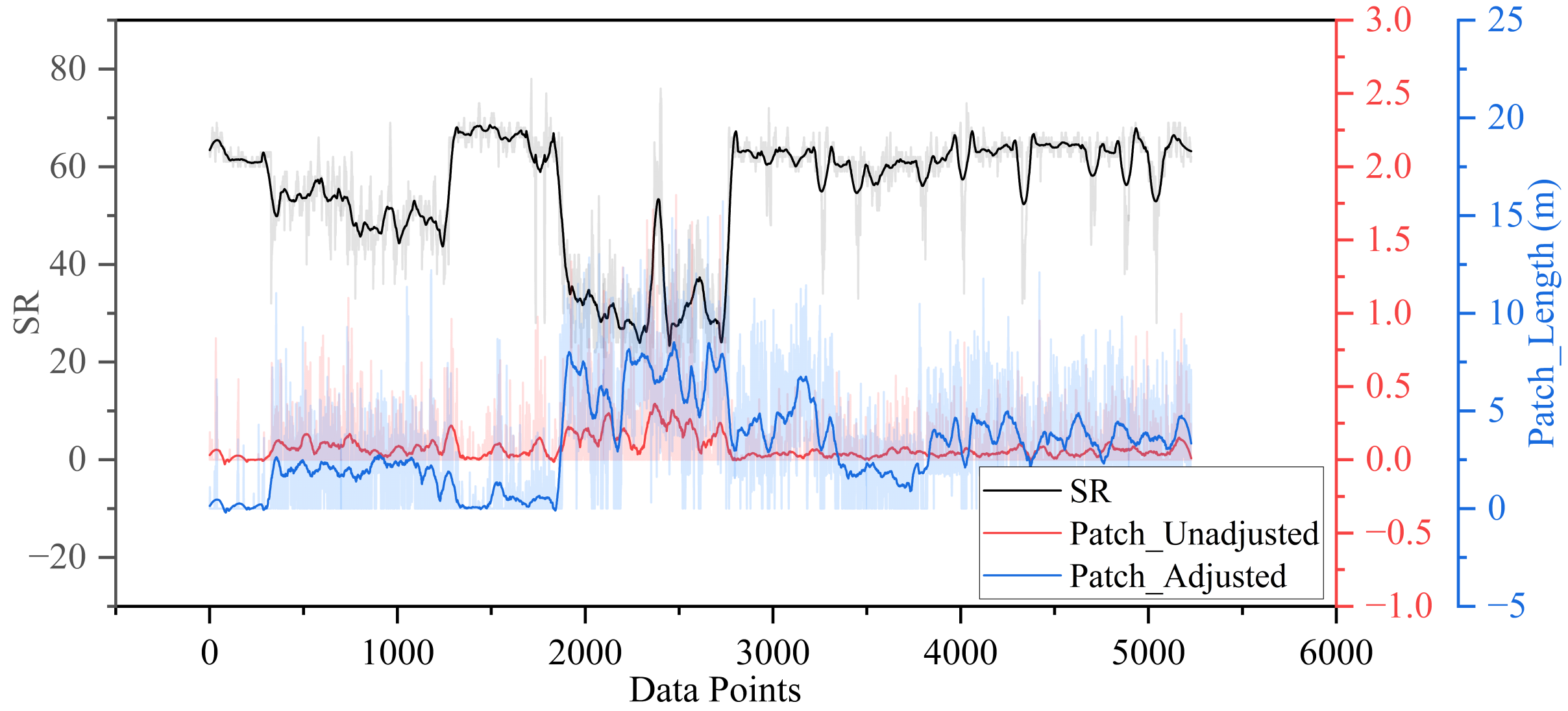}
\caption{Crack-patch length and SR in the test road}\label{fig:10}
\end{figure}

After detecting all the crack patches, the crack-patch length of each
image was automatically measured by the developed algorithm. Figure \ref{fig:10}
presents the statistical analysis of all length measurements within the
test road. From the figure, it is evident that the fluctuation in crack
lengths is substantial, which to some extent reflects the diversity of
road conditions. Figure 10 also illustrates the variation of SCRIM
Reading (SR), which reflects the pavement friction, at each detection
point. It is apparent from the figure that there is a correlation
between SR and crack patch length. As the crack length increases, there
is a corresponding decreasing trend in SR. Pearson correlation analysis
indicated a statistically significant moderate negative correlation
between the two variables, \(r = - 0.411,p < 0.001\). Spearman's rank
correlation analysis produced a consistent result,
\(\rho = - 0.336,p < 0.001\).

Furthermore, the patch length that was obtained from unadjusted images
is also demonstrated in Figure \ref{fig:10}. Williams' test indicated that the
correlation between \(SR\) and
\(Adjusted\ Patch\ Length\ \)(\(r = - 0.411\)) was significantly
stronger than the correlation between \(SR\)and
\(Unadjusted\ Patch\ Length\)(\(r = - 0.272\)), \(t(5225) = - 8.90\),
\(p < 0.001\).

\section{Conclusions}\label{conclusions}

This study developed an image-based framework for quantitatively
measuring pavement distress from front-view vehicle camera data without
requiring conventional field camera calibration. The proposed framework
integrates lane detection, perspective transformation, image-scale
adjustment, and object detection to convert pavement distress
measurements from pixel-based outputs into physically meaningful
dimensions. A homography matrix was constructed using lane boundaries to
transform front-view images into a vertical perspective, while standard
pavement markings were used as scale references to correct image
dimensions. Because the camera installation angle remains fixed for a
given vehicle, the perspective and scale parameters only need to be
determined once and can subsequently be applied to all images collected
by the same camera, thereby reducing manual processing requirements.

A refined training strategy was further adopted to improve the
adaptability of YOLO models to project-specific pavement images. The
results showed that refined training substantially improved the
detection performance of all evaluated models. Although YOLOv11m
achieved the highest overall accuracy, YOLOv11s was selected as the
final model because it provided a favorable balance between detection
accuracy and computational efficiency. The refined YOLOv11s achieved a
precision of 0.897, recall of 0.813, and mAP(0.5) of 0.881, with an
inference time of approximately 3.1 ms per image. The model also
demonstrated effective detection performance on both asphalt and
concrete pavement surfaces.

The detected bounding boxes were used to estimate the total crack-patch
length in each image after perspective and scale correction. The
resulting measurements exhibited substantial spatial variation along the
test roadway, reflecting differences in pavement conditions. In
addition, both Pearson and Spearman correlation analyses indicated a
statistically significant moderate negative relationship between
crack-patch length and SCRIM readings, suggesting that increased
crack-patch distress was generally associated with reduced pavement
friction. These findings demonstrate that quantitative distress
measurements extracted from vehicle images can provide
engineering-relevant information and can potentially support integrated
analyses of pavement surface condition and functional performance.

The proposed method provides a practical and computationally efficient
approach for processing large volumes of uncalibrated images collected
by conventional or connected vehicles. However, the current study relies
on visible and standardized pavement markings for scale correction and
assumes that the camera position and orientation remain unchanged. In
addition, the diagonal length of the bounding box represents an
approximation of crack-patch length rather than the exact distress
geometry. Future research should validate the measurement accuracy using
field measurements, investigate automatic calibration methods for roads
with missing or deteriorated markings, evaluate the influence of camera
vibration and installation changes, and extend the framework to other
distress types, including potholes, longitudinal cracks, and irregularly
shaped patches. These improvements could further support the large-scale
application of crowdsourced vehicle imagery in pavement condition
monitoring and asset management.

\nocite{PaddlePaddleND}
\bibliographystyle{plainnat}
\bibliography{references}

\end{document}